\documentclass[conference]{IEEEtran}
\IEEEoverridecommandlockouts
\usepackage{cite}
\usepackage{amsmath,amssymb,amsfonts}

\usepackage{algorithmic}
\usepackage{float}
\usepackage{graphicx}
\graphicspath{ {./images/} }
\usepackage{textcomp}
\usepackage{xcolor}
\usepackage[ruled,vlined]{algorithm2e}
\usepackage{subfig}
\def\BibTeX{{\rm B\kern-.05em{\sc i\kern-.025em b}\kern-.08em
    T\kern-.1667em\lower.7ex\hbox{E}\kern-.125emX}}
\begin{document}

\bstctlcite{IEEEexample:BSTcontrol}

\title{MAIDCRL: Semi-centralized Multi-Agent Influence Dense-CNN Reinforcement Learning}

\author{\IEEEauthorblockN{Ayesha Siddika Nipu}
\IEEEauthorblockA{\textit{Department of Computer Science} \\
\textit{Missouri State University}\\
Springfield, U.S. \\
Nipu62@MissouriState.edu}
\and
\IEEEauthorblockN{Siming Liu}
\IEEEauthorblockA{\textit{Department of Computer Science} \\
\textit{Missouri State University}\\
Springfield, U.S. \\
SimingLiu@MissouriState.edu}

\and
\IEEEauthorblockN{Anthony Harris}
\IEEEauthorblockA{\textit{Department of Computer Science} \\
\textit{Missouri State University}\\
Springfield, U.S. \\
Anthony999@MissouriState.edu}
}


\maketitle 

\IEEEpubidadjcol

\begin{abstract}
Distributed decision-making in multi-agent systems presents difficult challenges for interactive behavior learning in both cooperative and competitive systems. To mitigate this complexity, MAIDRL presents a semi-centralized Dense Reinforcement Learning algorithm enhanced by agent influence maps (AIMs), for learning effective multi-agent control on StarCraft Multi-Agent Challenge (SMAC) scenarios. In this paper, we extend the DenseNet in MAIDRL and introduce semi-centralized Multi-Agent Dense-CNN Reinforcement Learning, MAIDCRL, by incorporating convolutional layers into the deep model architecture, and evaluate the performance on both homogeneous and heterogeneous scenarios. The results show that the CNN-enabled MAIDCRL significantly improved the learning performance and achieved a faster learning rate compared to the existing MAIDRL, especially on more complicated heterogeneous SMAC scenarios. We further investigate the stability and robustness of our model. The statistics reflect that our model not only achieves higher winning rate in all the given scenarios but also boosts the agent's learning process in fine-grained decision-making.
\end{abstract}


\section{Introduction}\label{lab-intro}
Artificial Intelligence (AI) has advanced significantly in many aspects of our lives in recent years. The rapid progress in AI has reached the human-level or even outperformed human champions in a wide variety of tasks including autonomous driving, game playing, protein folding, and robotics. However, most of these achievements of AI are limited in single-agent systems where interaction among agents is not considered. Since there are a large number of applications that involve cooperation and competition between multiple agents, we are interested in AI techniques that work not only on single-agent systems but also multi-agent systems (MAS). 
Recently, deep Reinforcement Learning (DRL) has been considered for some of the most effective AI techniques to solve problems in many domains, i.e. \textit{AlphaGo} and \textit{AlphaStar} \cite{jumper2021highly}.
Extending DRL to enable interaction and communication among agents is critical to building artificially intelligent systems in multi-agent environments.
One of the main challenge of multi-agent RL (MARL) is that the canonical RL algorithms including Q-Learning and policy gradient algorithms do not generalize well to MAS due to the exponential growth of the number of states as the number of agents increases.
The second challenge in MARL is that the stationary Markovian property from an individual agent's perspective no longer exists in MAS due to the dynamic activities of other agents. These non-stationary states lead to significant stability issues to MARL in the learning process.
In addition, MAS themselves introduce extra cooperative and competitive learning tasks to achieve team objectives for individual agent decision-making.
A popular approach is to provide MARL with complete information for training agents, commonly named centralized learning. \\
Our prior work introduced agent influence maps (AIM), aggregated into a global multi-agent influence map (MAIM), which is then used in addition to local agent observations for fine-grained decision-making.
We combined MAIRL and the DenseNet model architecture, which defined Multi-Agent Influence Dense Reinforcement Learning (MAIDRL), and we evaluated MAIDRL's performance on StarCraft Multi-Agent Challenge (SMAC) scenarios in a real-time strategy (RTS) game, StarCraft II (SC2)\cite{harris2021maidrl}.
By extracting a descriptive representation from the complete global information and combining it with the DenseNet architecture, MAIDRL demonstrated a significant improvement in centralized, decentralized, and hybridized methods. 
In this study, we extended MAIDRL with the use of a Convolutional Neural Network (CNN) and introduce Multi-Agent Dense-CNN Reinforcement Learning (MAIDCRL) for solving MAS problems.
This reformulation of extracting spatial features from MAIM by utilizing multiple CNN layers further improved the learning performance of MAIDCRL in a variety of SMAC scenarios. 
In order to evaluate how the influence map (IM) affects the multi-agent learning performance of MAIDCRL, we therefore performed a rigorous analysis of the agent's behavior and found several fascinating behavioral attributes determined by the agent in the testing scenarios. 

\begin{figure*}[h]
    \centering
    \includegraphics[width=0.86\textwidth]{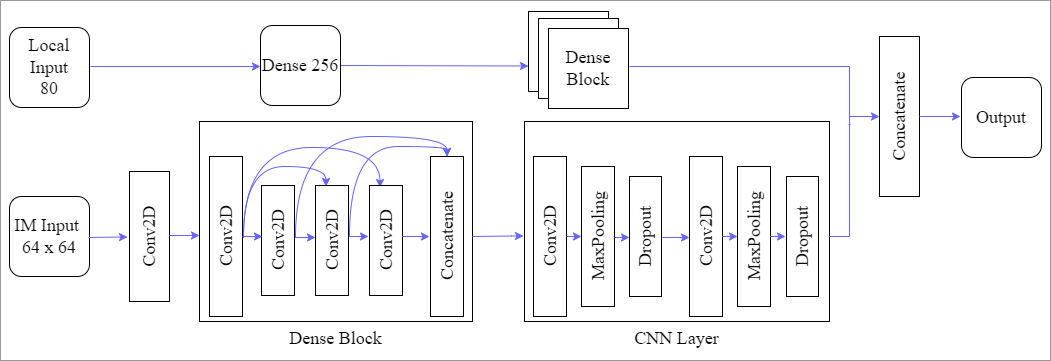}
    \caption{Outline of MAIDCRL architecture.}
    \label{fig:maidrl}
    \vspace{-4mm}
\end{figure*}

\section{Related Work}\label{relWork}
Extensive research has been conducted on applying different variants of RL algorithms for controlling agents in a wide variety of cooperative, competitive, and mixed MAS.
Nair et al. introduced parallel learning with a distributed network and a shared memory replay to split the learning tasks across multiple instances of simulations, effectively increasing the exploration speed at which agents learn \cite{Nair2015}. 
Liu et al. modeled the problem of MARL for a constrained, partially observable MDP, where the agents need to maximize a global reward function subject to both peak and average constraints. They proposed a novel algorithm, CMIX, to enable centralized training and decentralized execution under those constraints \cite{Liu2013, liu2013using}. 
Rashid et al. introduced QMIX, a network for training decentralized policies based on joint-action properties, which improved the performance of SC2 significantly \cite{rashid2018qmix}. 
There have also been several research on integrating Convolutional Neural Networks into DRL applications.
Stanescu et al. presented a CNN for RTS game state evaluation that goes beyond commonly used material-based evaluations by also taking spatial relations between units into account \cite{stanescu2016evaluating}. A Deep convolutional neural network (DCNN) has been developed by Kondo and Matsuzaki \cite{kondo2019playing} where they adopted supervised learning with multiple layers of convolutional networks and found exceptional scores. Other notable applications which enabled us to further investigate the broad spectrum of CNN are checking electricity prices, multi-microgrid co-operative systems, mimicking $Go$ Experts \cite{sutskever2008mimicking} etc. In the above research, they mainly focused on the application of CNN whereas, in our experiment, we are addressing the state representation and global information abstraction through the combination of MAIM and DCNN to provide shared goals and encourage collaborative behavior learning among agents.

\section{Methodology}\label{methodology}
In this research, we have evaluated the multi-agent learning performance of CNN-enabled MAIDCRL on SMAC platform. 
We conducted our experiments in selected homogeneous scenarios including \textit{3m}, \textit{8m}, \textit{25m} where the numeric value denotes the number of active marines in each team, and one heterogeneous scenario \textit{2s3z} in which two stalkers and three zealots work together to defeat the equal number of opponents at the beginning of each game. 
The SMAC game episodes are usually modeled as a Markov game, which is a multi-agent extension of MDPs \cite{Peng2017}.
A Markov game containing $N$ agents comprises a set of states $S$ that shows the status of the agents and the environment, and a set of actions $A_1,A_2,...,A_N$ and observations $O_1,O_2,...,O_N$ for each of the $N$ agents.
For an individual agent in the Markov game, we model the surrounding friendly and enemy units as a part of the environment and perform an action based on the agent's observation.
$S\times \{A_1, ..., A_N\}\rightarrow S',$ $r$ denotes the state transition from $S$ to $S'$ in a Markov game where each agent performs action following a policy $\pi$ in each environmental step and receives a shared reward $r$.
The theoretical maximum reward in each episode is scaled to a non-negative value between 0 to 20, as defined in SMAC.
The reward is a shared reward to the whole team instead of individual agents considering damage dealt to the enemy, points on a unit kill, and a bonus for victory.

\subsection{Experimental Features}
Our experiments are performed using Python, NumPy, and Tensorflow as the framework for all the RL training and evaluation. For statistical results, we evaluate the CNN-enabled MAIDCRL on $31$ random seeds in different SMAC scenarios. Each experiment runs for a total number of $1600$ episodes considering a large number of tuning parameters introduced in CNN architecture. MAIDCRL is built on top of the standard Advantage Actor-Critic (A2C) algorithm with separate neural network (NN) models for both actor and critic. 
We introduced a hybrid $\varepsilon$-greedy, softmax approach called $\varepsilon$-soft that starts with an initial value ($\varepsilon_0$) of $1$ and decreases throughout the overall running process. 

\begin{figure}[tb]
    \centering
        \subfloat[Average of the Running Average Episode Reward on $25m$.\label{SimpleAvg25m}]{\includegraphics[width=0.45\textwidth, height=4cm]{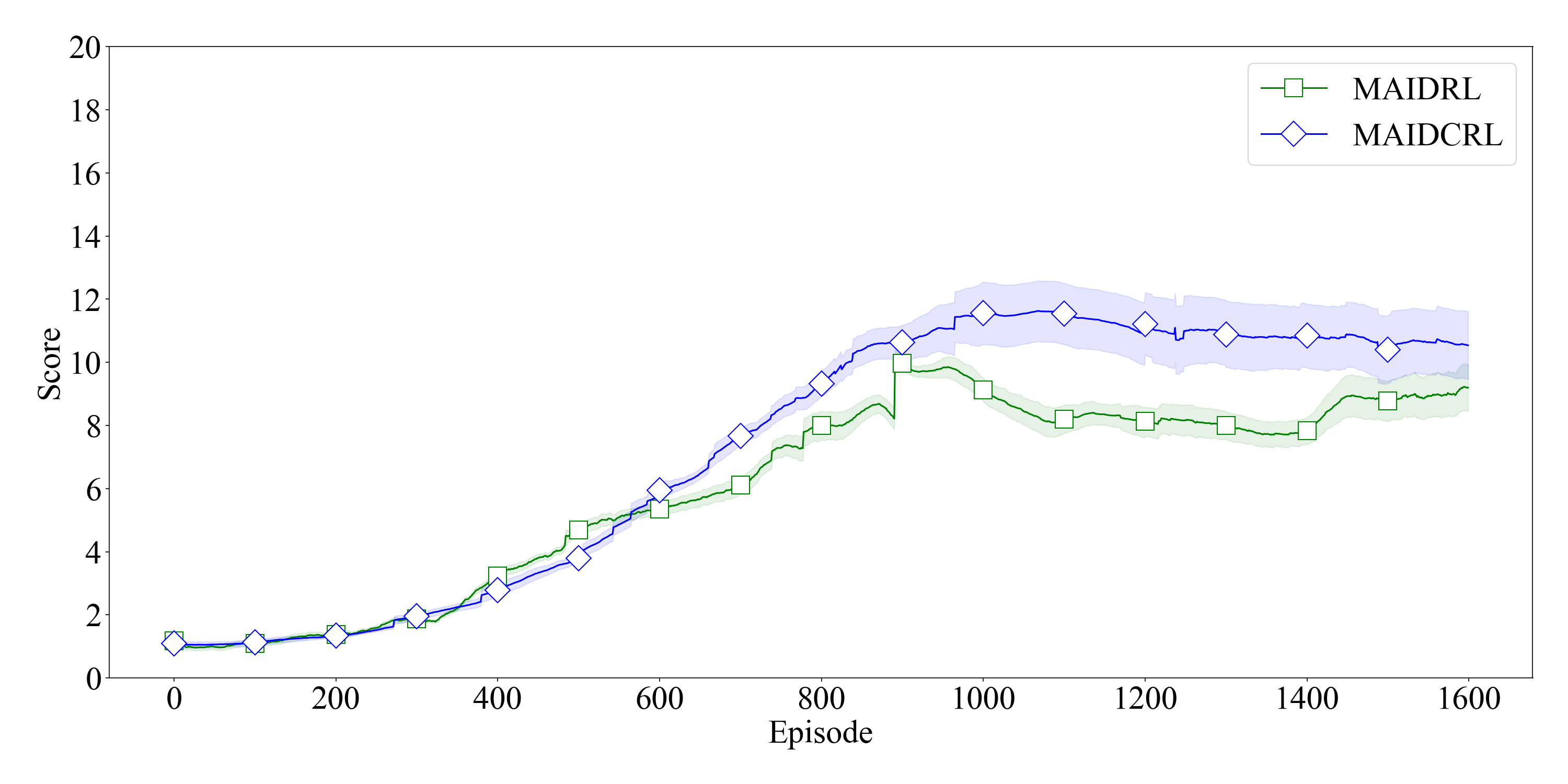}} \\
        \subfloat[Average of the Running Average Episode Reward on $2s3z$.\label{SimpleAvg2s3z}]{\includegraphics[width=0.45\textwidth, height = 4cm]{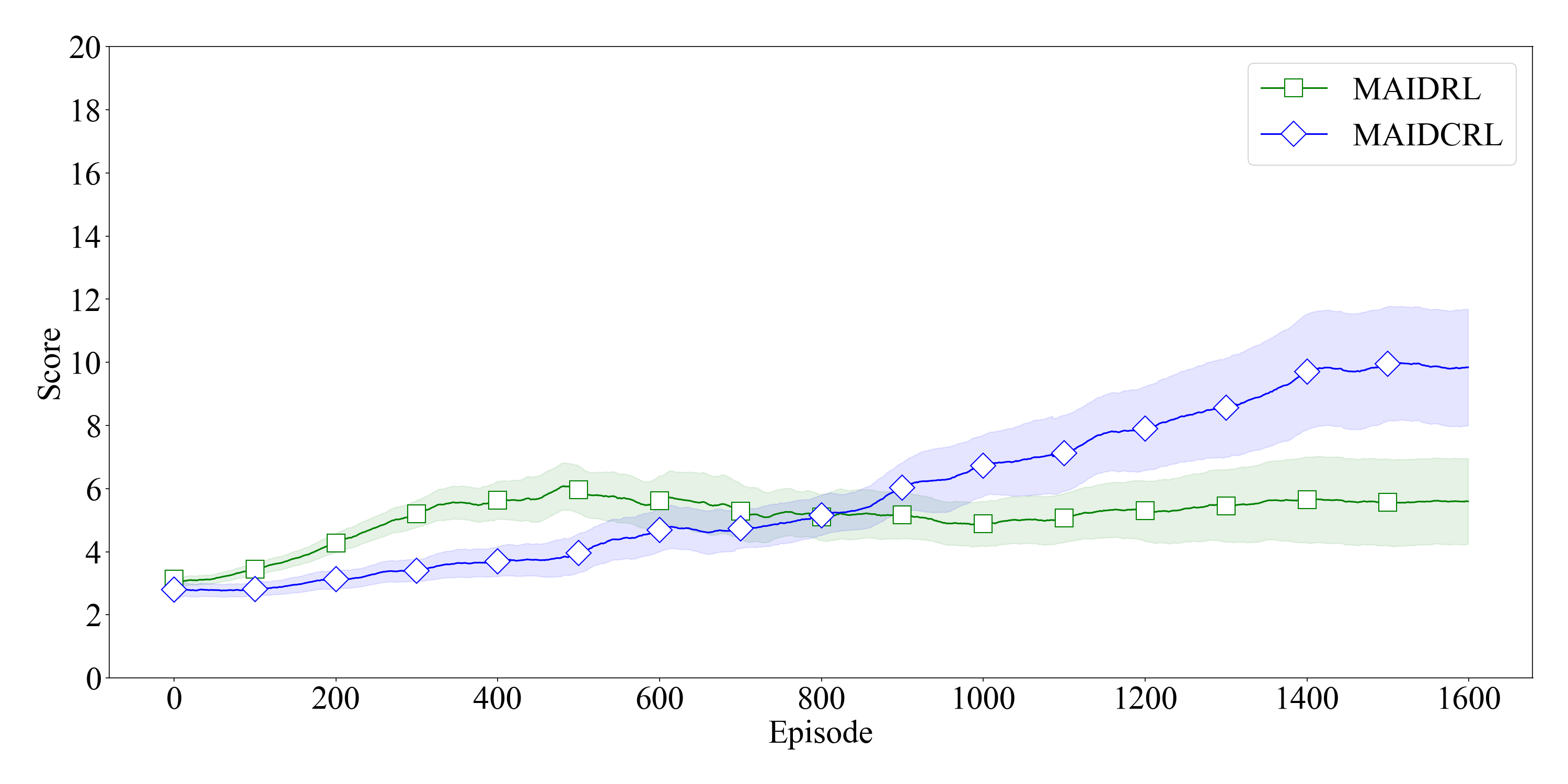}}
        
    \caption{Results of MAIDCRL on different scenarios.}
    \label{MAIDCRLStatGraph}
\end{figure}

\subsection{Multi-Agent Influence Dense-CNN Reinforcement Learning}
In our A2C RL configuration, we utilize two separate NN components without sharing neural layers between the actor and the critic networks. Each network contains one of the DenseNet grouping layers with three 256-neuron dense layers in the dense block.
Our A2C design contains one single controller that manages each agent individually based on the agent's observation, and the resulting NN allow for faster learning because the parameters in the neural network are updated based upon the parallel exploration of each agent.
At the terminal state of each episode, the parameters $\theta_A$ of the actor network are updated using the gradient ascent equation as shown in Equation \ref{gradientAscentEq} \cite{Mnih2016}.
\begin{equation}
    \nabla_{\theta_A}log\ \pi(a_t|s_t)\left(Q^\pi(a_t,s_t;\theta_A)-V^\pi(s_t;\theta_C)\right)
    \label{gradientAscentEq}
\end{equation}

In order to provide a shared goal and encourage collaborative behavior learning among agents, we use a spatial information technique, Agent Influence Map (AIM), to extract and filter useful features from the local information of each agent.
We aggregate the AIMs from all the agents on the map and generate a Multi-Agent Influence Map (MAIM), and we selected the dimensionality of $64\times64$ as it outperformed other dimensions on MAIDCRL.
We extended the existing MAIDRL model architecture by incorporating multiple convolutional layers that accept the MAIM as an input. 
The input of the influence map is concatenated with a CNN layer for each of the three groups in the dense blocks. 
Multiple convolutional layers have been incorporated in the new architecture with 32 filters for each layer, a stride rate of 1, and a kernel size of 3 to extract spatial features on MAIM. 
We also used $elu$ activation function, a max-pooling of $2 \times 2$, and different dropouts ranges [0.1, 0.5]. We explored and compared the outcome of MAIDCRL representation with MAIDRL and collected statistics for several different combinations. Fig. \ref{fig:maidrl} demonstrates the detailed architecture of our proposed MAIDCRL model.

\begin{table}[tb]
\caption{Performance Comparison between MAIDCRL and MAIDRL on Extended Scenarios}
\begin{center}
\begin{tabular}{c c c c c c}
\hline
\textbf{Scenario} & \textbf{Method} & \textbf{Min} & \textbf{Max} & \textbf{Avg} & \textbf{Std}\\
\hline
$3m$ & MAIDRL & 4.29 & \textbf{17.14} & 11.01 & 4.19 \\
& MAIDCRL & \textbf{5.01} & 16.98 & \textbf{14.84} & \textbf{3.12} \\
\hline
$8m$ & MAIDRL & 5.01 & 16.93 & 14.77 & \textbf{3.77} \\
& MAIDCRL & \textbf{6.82} & \textbf{18.79} & \textbf{15.65} & 3.81 \\
\hline
$25m$ & MAIDRL & 6.45 & 13.59 & 11.82 & \textbf{0.94} \\
& MAIDCRL & \textbf{9.22} & \textbf{16.09} & \textbf{12.95} & 1.25 \\
\hline
$2s3z$ & MAIDRL & \textbf{5.32} & 13.70 & 10.33 & \textbf{3.43} \\
& MAIDCRL & 5.12 & \textbf{18.88} & \textbf{12.14} & \textbf{2.25} \\
\hline
\end{tabular}
\label{peakResultCNNStats}
\end{center}
\vspace{-4mm}
\end{table}

\begin{figure}[tb]
    \centering
        \subfloat[Total Number of Winning Among All Seeds \label{TotalWinning}]{\includegraphics[width=0.45\textwidth, height=4cm]{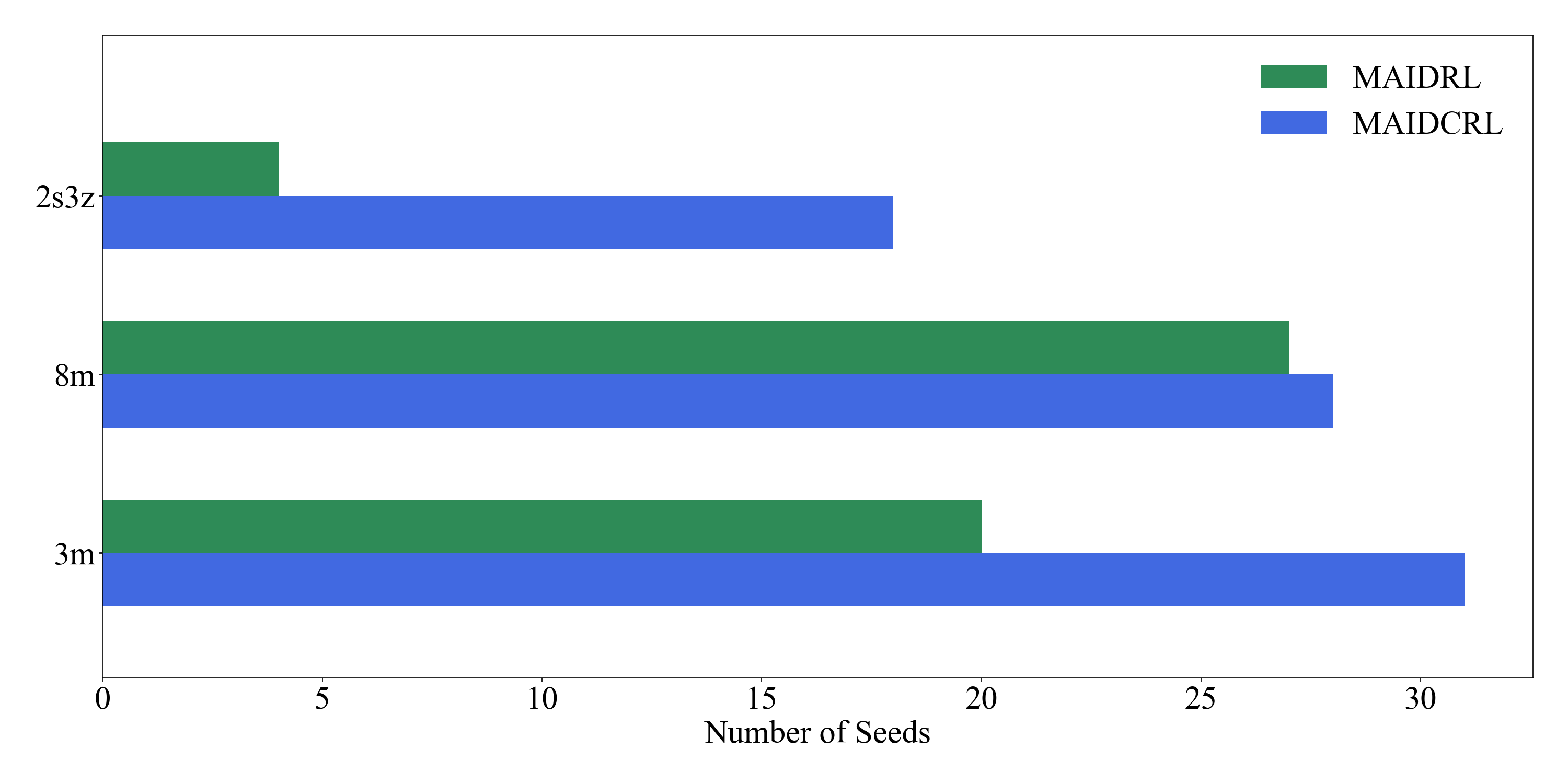}} \\
        \subfloat[Average Number of Episodes for First Winning\label{FirstWinning}]{\includegraphics[width=0.45\textwidth, height=4cm]{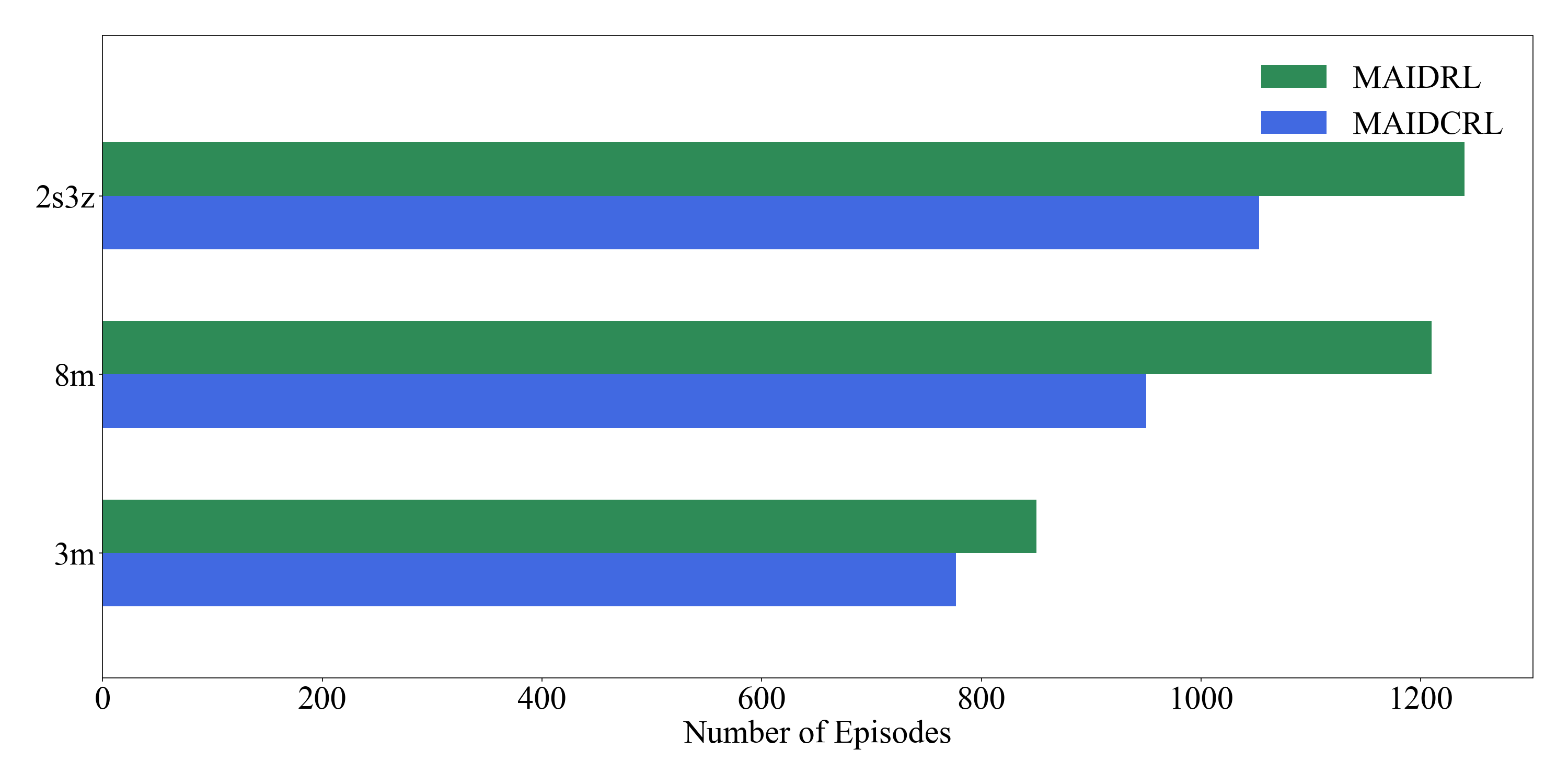}}
    \caption{Robustness of MAIDCRL}
    \label{robustness}
    \vspace{-2mm}
\end{figure}

\section{Results and Discussion}\label{results}
We evaluated the MAIDCRL performance from three different aspects: the average running episode reward, the average of total number of winnings, and the average number of episodes for achieving the first win. These criteria are used to evaluate the overall performance, overall robustness, and the speed of the learning algorithms respectively. 

\subsection{MAIDCRL Learning Performance}
Fig.~\ref{MAIDCRLStatGraph} shows the results received from MAIDCRL and MAIDRL on complex $25m$ and $2s3z$ scenarios. 
The maximum running average is improved by $18.39\%$ on $25m$ and $37.81\%$ on $2s3z$.
Fig.~\ref{SimpleAvg25m} and Fig.~\ref{SimpleAvg2s3z} illustrate that the learning is slow at the beginning for complicated SMAC scenarios. However, MAIDCRL outperformed MAIDRL after $600$ and $800$ episodes on average for $25m$ and $2s3z$ respectively.   

Table~\ref{peakResultCNNStats} shows the detailed results of MAIDCRL and MAIDRL comparison on all scenarios including $3m$, $8m$, $25m$ and $2s3z$.
Note that the boldly marked values indicate the best performance in the given scenarios. 
Our CNN-enabled MAIDCRL achieved a higher running average score over all the testing scenarios. MAIDRL seems to achieve maximum reward on $3m$ scenario, whereas MAIDCRL surpassed by $18.79$, $16.09$, and $18.88$ on $8m$, $25m$, and $2s3z$ respectively.

\begin{figure*}[h]
    \centering
    \includegraphics[width=1.0\textwidth]{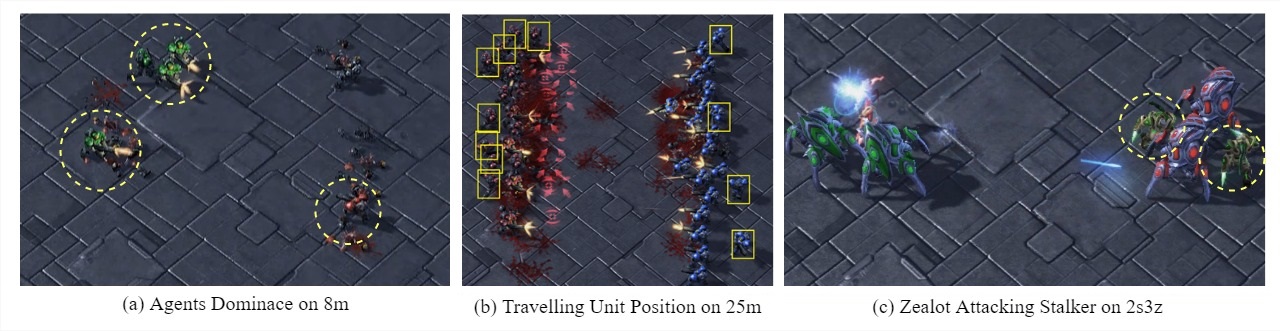}
    \caption{Learned Behavior on Different SMAC Scenarios}
    \label{learnedBehavior}
    \vspace{-4mm}
\end{figure*}

\subsection{Robustness of MAIDCRL Architecture}
To evaluate the robustness of the presented approach, we further analyzed the total number of winning instances over the total of 31 seeds of both MAIDCRL and MAIDRL on scenarios $3m$, $8m$ and $2s3z$. Figure~\ref{TotalWinning} illustrates that MAIDCRL won all runs on $3m$, 28 on $8m$, and 18 on $2s3z$ whereas MAIDRL couldn't beat our proposed model in any of the challenging scenarios.
We further had a close observation on the learning speed of our solution. Figure~\ref{FirstWinning} shows that CNN-enabled MAIDCRL found the first winning strategy after $770$, $950$, and $1053$ episodes on average for $3m$, $8m$ and $2s3z$ scenarios respectively.
It takes $800$, $1200$, and $1240$ episodes for MAIDRL to win on $3m$, $8m$, and $25m$ respectively. Therefore, CNN-enable MAIDCRL learned faster compared to MAIDRL in all the given homogeneous and heterogeneous scenarios. 
Note that none of these two models achieved a winning strategy in $25m$, thus we have skipped $25m$ scenario from this evaluation.

\subsection{Learned Behavior Analysis}

A qualitative comparison of learned behaviors is also analyzed to the best of our interest. The best performing RL models on each scenario are selected as the trained controllers in the test runs.
While observing the episodes played by the MAIDCRL controller, we have noticed two major strategies, one is prioritizing collaborative attack and another one is repositioning with minimal movement after damage. 
The behavior specific characteristics and their impact is illustrated in the following subsections. 
When we loaded our pre-trained RL model on $8m$ scenario, a collective movement was noticed among the agents even though all the agents made their decisions in a completely decentralized way. Therefore they were more successful to win than randomly selecting their target units. Figure~\ref{learnedBehavior}a depicts the dominance of the MAIDCRL model over the SC2 AI agents where four controlled agents remained alive and worked in two groups targeting only one enemy unit.
From our observation of $25m$, we found that some of the frontliners adjusted their position in a way so that they can reorganize their position with minimal movement in case of healing damages. Figure~\ref{learnedBehavior}b reflects that the agents are more reluctant than SC2 AI while positioning each of the units. As a result, they actively focus on dealing with the damages of opponent's attack. 
On the heterogeneous scenario $2s3z$, we observed that prioritizing the target played a significant role while there are multiple types of units in the enemy team. Figure~\ref{learnedBehavior}c shows that our melee unit zealots moved passed enemy zealots in the front line and focused fire on enemy stalkers in the back first.

\section{Conclusion and Future Work}\label{conclusion}
In this study, we extended a semi-centralized RL model MAIDRL and introduced a new CNN-enabled MAIDCRL to solve various MARL systems. We evaluate the performance of MAIDCRL in homogeneous and heterogeneous SMAC scenarios of varying complexity for the statistical results.  
MAIDCRL demonstrated observable improvements in the overall performance, overall robustness, generalizability, and peak performance shown in selected scenarios. These comparisons could lead to new ideas on how to design better influence maps and discover new features in Multi-Agent reinforcement learning algorithms. Further investigation is required to test our model in a wider range of heterogeneous environments containing more complicated maps.

\bibliographystyle{IEEEtran}
\bibliography{Ayesha}

\end{document}